\documentclass[11pt]{article}
\usepackage{booktabs}
\usepackage{amsmath}
\usepackage{amssymb}

\usepackage[final]{acl}

\usepackage{times}
\usepackage{latexsym}
\usepackage[table]{xcolor}

\usepackage[T1]{fontenc}
\usepackage[utf8]{inputenc}

\usepackage{microtype}

\usepackage{inconsolata}

\usepackage{graphicx}
\usepackage{multirow}
\definecolor{bestblue}{RGB}{0,0,255}

\newcommand{\dd}[2]{#1\,(#2)}
\newcommand{\ddB}[2]{#1\,\textcolor{bestblue}{(#2)}}
\definecolor{refgray}{gray}{0.93}

\usepackage[normalem]{ulem}

\title{Child ASR Adaptation with Adult Retention: An Empirical Study}

\author{
 \textbf{Houssam Eddine-Othman Lachemat},
 \textbf{Shammur Absar Chowdhury}
\\
 Qatar Computing Research Institute,
\\
 \texttt{\{hlachemat, shchowdhury\}@hbku.edu.qa}
}

\begin{document}
\maketitle
\begin{abstract}
Automatic Speech Recognition (ASR) systems often underperform for children and non-native speakers, while adapting adult ASR models to child speech can cause adult-speech forgetting. We study \emph{child ASR adaptation with adult retention} across Arabic and English. We compare full fine-tuning, LoRA, and post-hoc weight-space merging across encoder--decoder, encoder--CTC, and AudioLLM-based ASR systems. Experiments use Arabic native and non-native child speech, English MyST child speech, and adult benchmarks from MGB-2 and LibriSpeech test-clean. We evaluate recognition quality with WER and quantify the adaptation--retention trade-off using Retention Index, Child Adaptation Gain, and Adaptation Recovery. Results show that child adaptation is necessary, especially for non-native Arabic and English child speech, but direct adaptation often reduces adult ASR performance. Bilingual adaptation is more stable than language-specific adaptation. Weight-space merging often improves the trade-off, especially for encoder--CTC, Whisper, and AudioLLM-based ASR, with LERP favoring adult retention and TIES recovering stronger child gains. For the encoder--decoder model, direct bilingual fine-tuning remains strongest in raw WER.\footnote{Code, and models are available at \url{https://github.com/qcri/Child-ASR-Adaptation}.} 
\end{abstract}

\section{Introduction}
Automatic Speech Recognition (ASR) has achieved strong performance on adult speech, but children's speech remains difficult to recognize. Compared with adults, children often have higher pitch, shorter vocal tracts, greater pronunciation variability, and more disfluencies, all of which can degrade ASR accuracy \cite{wu2019childasr,app12094419}. The problem is further amplified by the limited availability of labeled child-speech data and by bilingual settings, where models must handle both age-related and language-related shifts.

A common solution is to adapt adult-trained ASR models using child-speech data. This can improve child recognition on datasets such as MyST \cite{pradhan-etal-2024} and CMU Kids \cite{ldc_cmu_kids_1997}, but it can also reduce performance on adult speech due to catastrophic forgetting. This creates a deployment challenge for systems that must serve both adults and children, such as home assistants, classroom tools, and bilingual speech interfaces. We therefore study the \emph{adaptation--retention} trade-off: improving child ASR while preserving adult ASR performance.

\begin{figure}
    \centering
    \includegraphics[width=1.0\linewidth]{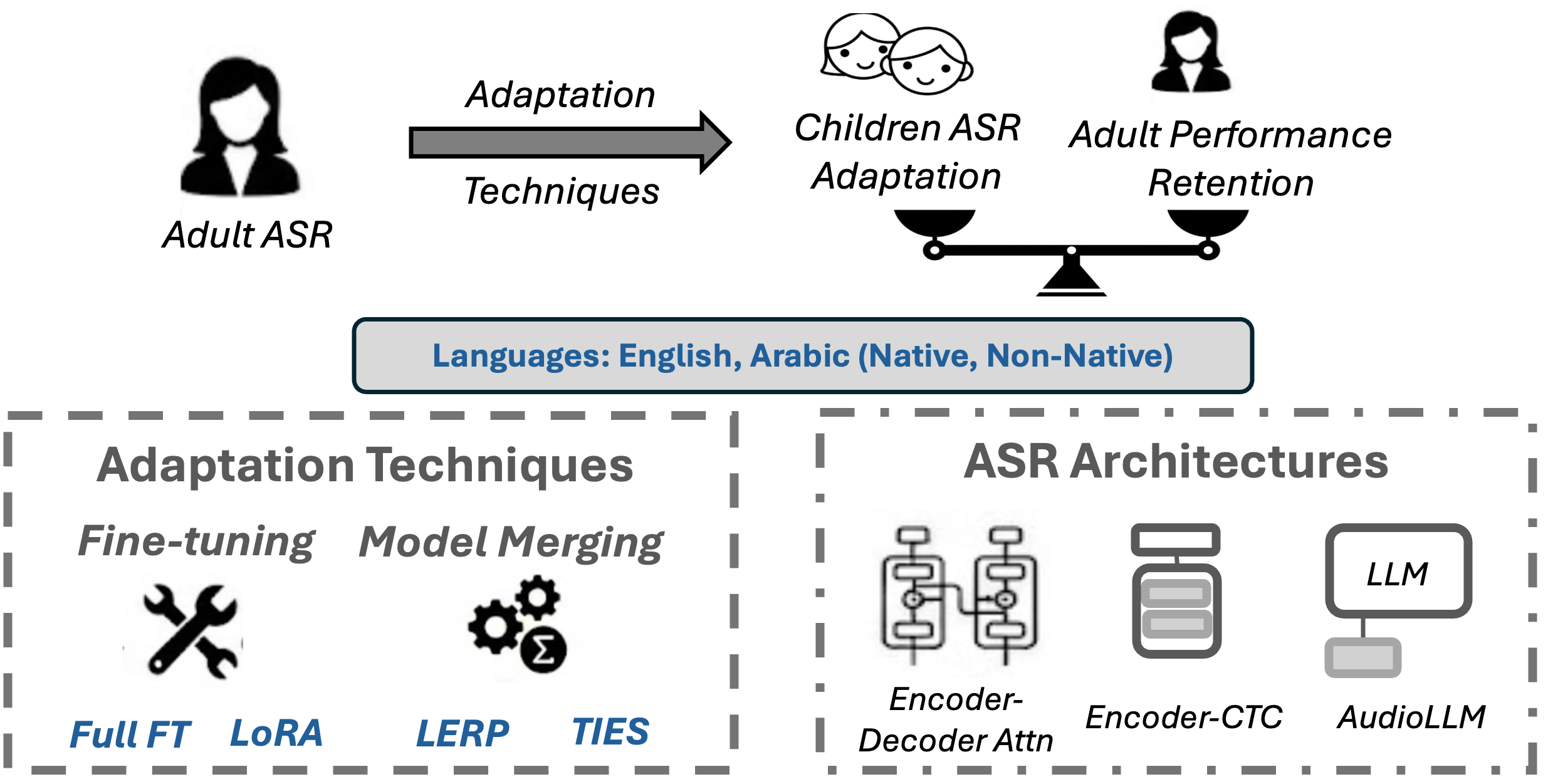}
    \caption{Overview of the child ASR adaptation pipeline across architectures, adaptation methods, and adult-retention evaluation.}
    \label{fig:overview}
\end{figure}

This paper presents a retention-aware empirical study of adult-to-child ASR adaptation. We compare three practical adaptation strategies: full fine-tuning, LoRA-based parameter-efficient adaptation \cite{hu2021lora}, and post-hoc weight-space merging. For merging, we evaluate Linear Interpolation (LERP) and TIES \cite{yadav2023ties}, following the broader motivation of model soups and task arithmetic \cite{wortsman2022modelsoups,ilharco2022taskarithmetic}. Unlike fine-tuning, merging combines adult and child-adapted checkpoints directly in parameter space without additional supervised training, allowing us to test whether child gains can be recovered while reducing adult forgetting. Figure~\ref{fig:overview} summarizes our experimental pipeline.

We evaluate these strategies across four ASR families: a bilingual encoder--decoder model, a multilingual encoder--decoder model (Whisper Large v3 \cite{pmlr-v202-radford23a}), a bilingual encoder--CTC model,  and a multilingual AudioLLM-based ASR system built from Qwen2.5--Omni--7B and adapted for Arabic--English ASR. Our evaluation covers Arabic native and non-native child speech from AraYoungVoices \cite{chowdhury2026arayoungvoices} (adaptation on ages 7--18; evaluation on its AraKids 7--12 test split), English child speech from MyST \cite{pradhan-etal-2024}, and adult retention benchmarks from LibriSpeech test-clean \cite{panayotov2015librispeech} and MGB-2 \cite{ali2016mgb}. In addition to WER, we report Retention Index, Child Adaptation Gain, and Adaptation Recovery to quantify both recognition improvement and retention cost.

Our results reveal several consistent trends. First, child adaptation is necessary across architectures, especially for non-native Arabic children and for English child speech in bilingual settings. Second, language-specific adaptation can cause cross-language degradation, while bilingual adaptation is generally more stable. Third, direct fine-tuning often gives strong child performance gains. However, it can substantially reduce adult ASR performance, particularly on MGB-2. Fourth, weight-space merging often improves the adaptation--retention balance, especially for encoder--CTC, Whisper, and AudioLLM-based ASR systems. LERP tends to preserve adult performance, while TIES more aggressively recovers child-speech gains. For the encoder--decoder model, however, direct bilingual fine-tuning remains strongest in raw WER.

Overall, this study provides practical guidance for building ASR systems that better serve both children and adults. It shows retention-aware evaluation is essential for child ASR adaptation and model merging can be a simple and effective tool for reducing the cost of adult-speech forgetting.

\section{Related Work and Positioning}% \label{sec}

\subsection{Child speech recognition and benchmarks}
ASR trained predominantly on adult speech degrades on children due to acoustic and linguistic mismatch and limited labeled child data. 
Recent child-ASR studies have explored diverse modeling approaches, including
discrete speech representations, end-to-end neural architectures, and adaptation
of large pretrained ASR models to children's speech
\cite{dutta2025exploring,vrunda2024,shivakumar2022end,jain2023adaptationwhispermodelschild}.
Adult$\rightarrow$child adaptation is a pragmatic route to lower child WER, but raises an often under-emphasized, deployment-critical question: how to improve child recognition \emph{without sacrificing adult competence} when one system serves both populations \cite{app12094419,shivakumar2020adult2child_transfer,jain2023adaptationwhispermodelschild, 10887889}.
Classic corpora (OGI Kids, PF-STAR) and modern benchmarks (CMU Kids read speech, MyST conversational child--agent dialogue) span controlled and conversational settings \cite{shobaki2000ogikids,pfstar2005,ldc_cmu_kids_1997,pradhan-etal-2024}.

Prior child-ASR research has largely focused on English, with comparatively
limited evaluation in other languages. We therefore extend the analysis to
Arabic child speech to examine whether adaptation gains and adult-speech
retention generalize across language and nativity shifts.
Bilingual and L2 settings add further difficulty, as age-related variability
interacts with cross-lingual phonology and accent transfer; PF-STAR includes
non-native child speech, while TLT-school targets L2 learners in school
settings \cite{pfstar2005,gretter2020tlt_school}. For Arabic, child-specific
resources remain scarce, and strong pretrained models still show substantial
gaps relative to adult speech \cite{lu2022ssl_child,alkadri2025arabiclittlestt}.
AraYoungVoices provides an Arabic child-adolescent-speech resource covering both native
(L1) and non-native (L2) speakers. In this work, we use AraKids (the children subset) as the
Arabic data for this task, enabling evaluation under combined age and nativity
shift.
% \rev{Prior child-ASR research has largely focused on English, with comparatively limited evaluation in other languages. We therefore extend the analysis to native and non-native Arabic child speech to assess whether adaptation gains and adult-speech retention patterns generalize across language and nativity shifts.}
% Bilingual and L2 settings compound the difficulty, as age-related variability interacts with cross-lingual phonology and accent transfer; PF-STAR includes non-native child speech and TLT-school targets non-native children in school contexts \cite{pfstar2005,gretter2020tlt_school}. For Arabic, scarce child-specific corpora remain a major barrier, and recent dataset efforts show that even strong pretrained models stay far from adult-level accuracy on Arabic child speech \cite{lu2022ssl_child,alkadri2025arabiclittlestt}. This gap motivated \revdel{AraKids}\rev{AraYoungVoices (ages 7--18)}, with native (L1) and non-native (L2) children speaking Arabic, as a benchmark for adaptation under combined age and nativity shift.

\subsection{Adult ASR adaptation to Children and the cost downside}
Child adaptation mainly targets end-to-end architectures, including encoder--CTC \cite{graves2006ctc} and attention-based encoder--decoder models \cite{chan2015las}, often with Conformer-style encoders \cite{gulati2020conformer}. Large-scale pretraining has produced strong general-purpose backbones (e.g., wav2vec~2.0 and Whisper) adaptable with modest labeled data \cite{baevski2020wav2vec2,pmlr-v202-radford23a}, and speech-language stacks pairing audio encoders with LLMs have emerged as alternative ASR backbones/decoders \cite{audiopalm2023,tang2024salmonn,chu2024qwen2audio}. These trends motivate comparing adaptation strategies across \emph{multiple} model families rather than a single architecture.
A growing body of work fine-tunes pretrained ASR on child corpora (e.g., MyST, CMU Kids, PF-STAR), including Whisper adaptation studies and variants that control data quality or mix adult/childrenized speech to reduce mismatch \cite{jain2023adaptationwhispermodelschild,attia2024kidwhisper,southwell2024classroom_asr,graave2024mixed_finetuning,10887889}. These studies consistently show large child-domain gains, but evaluation is frequently centered on child-only WER, leaving the \emph{adult regression} induced by child adaptation less systematically quantified. When one system must serve both adults and children, child-only WER can look like progress while masking a real adult regression that the same system still has to absorb. This motivates retention-aware evaluation where adult ASR performance is explicitly anchored as a first-class constraint.
Sequential adaptation induces catastrophic forgetting, where optimizing for a new domain degrades previously learned ones. Continual learning mitigates this via regularization (e.g., EWC), distillation-style constraints (e.g., Learning without Forgetting), and evaluation constructs such as forgetting/backward transfer computed from task-wise performance matrices \cite{kirkpatrick2017ewc,li2016lwf,lopezpaz2017gem,chaudhry2018riemannian_walk}. In speech recognition, continual learning and domain expansion studies similarly report old-domain degradation and explore constraints for speech models \cite{hadian2021lfmmi_cl}. Our study operationalizes this perspective with adult anchors and retention-focused metrics (ratio-based retention/recovery scores), enabling direct comparison of adaptation strategies under an explicit adult-performance requirement.

\subsection{Parameter-efficient adaptation and weight-space merging}
Parameter-efficient fine-tuning (PEFT) adapts large pretrained backbones while updating only a small number of parameters, reducing training and storage cost. Adapters and LoRA are widely used in NLP \cite{houlsby2019adapters,hu2021lora} and have recently been explored for adapting speech foundation models (e.g., Whisper) in multilingual settings \cite{song2024lora_whisper}. Related child-ASR work also studies lightweight modules and constrained updates to mitigate mismatch in self-supervised representations \cite{fan2022draft}. We include LoRA as a \emph{practitioner-standard} baseline and compare its child gains and adult regression against full fine-tuning and post-hoc weight-space merging under the same adult anchors and retention metrics across ASR architectures, framing the study as a controlled empirical comparison rather than a new adaptation method.
Weight-space model merging offers a complementary, post-hoc alternative: combining an adult anchor and a child specialist without additional gradient-based training. Model soups average across multiple fine-tuned checkpoints, while task arithmetic interprets fine-tuning as task vectors composed by addition/subtraction \cite{wortsman2022modelsoups,ilharco2022taskarithmetic}. TIES-Merging further addresses destructive interference by pruning small-magnitude updates, resolving sign conflicts, and selectively merging aligned parameters \cite{yadav2023ties}. These methods are attractive for \emph{adult-retentive} child adaptation because they can recover child gains while preserving adult performance relative to naive child-only fine-tuning.

\subsection{Positioning this work}
Building on prior child ASR adaptation studies and continual-learning insights on forgetting, we frame \emph{child ASR adaptation with adult retention} as a retention-aware empirical comparison rather than a new algorithmic proposal. We systematically evaluate (i) full fine-tuning, (ii) LoRA-based PEFT, and (iii) post-hoc weight-space merging (LERP and TIES) across encoder--decoder attention models, encoder--CTC models, Whisper, and an AudioLLM-based ASR. We anchor adult performance on LibriSpeech test-clean and MGB-2 and extend child evaluation beyond English with AraYoungVoices (the test set is Arabic L1/L2, 7--12 years-old age group, referenced to as AraKids) and MyST, enabling a direct characterization of adaptation--retention trade-offs across architectures and languages \cite{jain2023adaptationwhispermodelschild,10887889,fan2022draft,yadav2023ties,pradhan-etal-2024,pmlr-v202-radford23a}.
The contribution is therefore empirical rather than algorithmic. We study adult-to-child adaptation under combined age, language, and nativity shifts, and we show that the trade-off between child-speech improvement and adult-speech retention depends on the architecture.

\section{Datasets and Benchmarking}
\label{sec:data}

We adapt models using child-speech data and evaluate child WER on held-out child test sets. 
Adult-speech datasets only serve as retention anchors to measure how much adult ASR performance is preserved after child adaptation (Tables~\ref{tab:data_overview},  \ref{tab:arakids_overview}).

\begin{table}[t]
\centering
\setlength{\tabcolsep}{3pt}
\renewcommand{\arraystretch}{1.12}

\resizebox{\columnwidth}{!}{%
\begin{tabular}{l l c c c c}
\hline
\textbf{Domain} & \textbf{Dataset} & \textbf{Trn} & \textbf{Dev} & \textbf{Tst} & \textbf{Role} \\
\hline
Child & AraYoungVoices (Ar) & \checkmark & \checkmark & \checkmark & Adapt + child eval \\
% Child & AraKids-L2 & \checkmark & \checkmark & \checkmark & Adapt + child eval \\
Child & MyST (En)  & \checkmark & \checkmark & \checkmark & Adapt + child eval \\
\hline
Adult & MGB-2 (Ar)  & -- & \checkmark & \checkmark & Adult retention \\
Adult & LibriSpeech (En) & -- & \checkmark & \checkmark & Adult retention \\
\hline
\end{tabular}%
}
\caption{Datasets and split usage. Child data is used for adaptation and child evaluation, while adult data is used only for retention evaluation. For Arabic, training and development data come from the full
AraYoungVoices corpus (ages 7--18), while child evaluation uses the AraKids
USUP test set (ages 7--12).}
\label{tab:data_overview}
\end{table}

\begin{table}[!t]
\centering
\scriptsize
\setlength{\tabcolsep}{3pt}
\renewcommand{\arraystretch}{0.95}
\begin{tabular*}{\columnwidth}{@{\extracolsep{\fill}}lrrr@{}}
\toprule
\textbf{Dataset} & \textbf{\#Spk} & \textbf{\#Utt} & \textbf{Hours} \\
\midrule
\multicolumn{4}{@{}c}{\textit{Training data}} \\ \midrule
% \addlinespace[2pt]
AraYoungVoices total        & 199 & $\approx$62K  & 102.94 \\
\quad AraYoungVoices L1     & 107  & $\approx$32K  & 45.40 \\
\quad AraYoungVoices L2     & 92  & $\approx$329K  & 57.54 \\ 
MyST                 & --  & $\approx$77K  & 171.20 \\\midrule
% \addlinespace[1pt]
\textbf{Total}       & \textbf{--} & \textbf{$\approx$141K} & \textbf{280.38} \\
\bottomrule
\multicolumn{4}{@{}c}{\textit{Test data}} \\\midrule
% \addlinespace[2pt]
AraKids total        & 66  & 2K     & 4.00 \\
\quad AraKids L1     & 26  & 1K     & 1.57 \\
\quad AraKids L2     & 40  & 1K     & 2.43 \\
MyST                 & --  & 13,180 & 28.32 \\\midrule
% \addlinespace[1pt]
\textbf{Total}       & \textbf{--} & \textbf{15,180} & \textbf{32.32} \\
\bottomrule
\end{tabular*}
\caption{Child adaptation data overview. Training rows report the AraYoungVoices corpus (ages 7--18), split into native (L1) and non-native (L2) speakers; test rows report the AraKids USUP test set (ages 7--12), which is the Arabic child evaluation set used throughout the paper. For MyST, we use the official split.} 
\label{tab:arakids_overview}
\end{table}

\begin{table}[!t]
\centering
\scriptsize
\setlength{\tabcolsep}{3pt}
\renewcommand{\arraystretch}{0.95}
\begin{tabular*}{\columnwidth}{@{\extracolsep{\fill}}lrrrrr@{}}
\toprule
& \multicolumn{3}{c}{\textbf{Train}} & \multicolumn{2}{c}{\textbf{Test}} \\
\cmidrule(lr){2-4}\cmidrule(lr){5-6}
\textbf{Group} & \textbf{\#Spk} & \textbf{\#Utt} & \textbf{Hours} & \textbf{\#Spk} & \textbf{\#Utt} \\
\midrule
\multicolumn{6}{l}{\textit{L1 (native)}} \\
Egyptian        & 33 & 9{,}906  & 14.45 & 5  & 252 \\
Gulf            & 25 & 7{,}429  & 10.95 & 7  & 249 \\
Levantine       & 25 & 7{,}089  & 8.51  & 7  & 250 \\
North African   & 24 & 7{,}853  & 11.50 & 7  & 249 \\
\midrule
\multicolumn{6}{l}{\textit{L2 (non-native)}} \\
Africa          & 13 & 4{,}242  & 9.23  & 7  & 250 \\
Americas        & 14 & 4{,}582  & 10.01 & 5  & 250 \\
Asia            & 30 & 9{,}449  & 18.26 & 15 & 250 \\
Europe          & 35 & 11{,}024 & 20.04 & 13 & 250 \\
\midrule
\textbf{Total}  & \textbf{199} & \textbf{61{,}574} & \textbf{102.94} & \textbf{66} & \textbf{2{,}000} \\
\bottomrule
\end{tabular*}
\caption{AraYoungVoices distribution by group. L1 groups are Arabic macro-regions;L2 groups are the speakers' regions of origin. Train and test speakers are disjoint. Train columns report the AraYoungVoices training set (ages 7--18);
test columns report the AraKids USUP test set (ages 7--12).} 
\label{tab:arakids_train_regions}
\end{table}

% \begin{table}[!t]
% \centering
% \scalebox{0.63}{
% \begin{tabular}{lrrr}
% \toprule
% \textbf{Split} & \textbf{\#Spk} & \textbf{\#Utt} & \textbf{Hours} \\
% \midrule
% \multicolumn{4}{l}{\textit{Training data}} \\
% \midrule
% AraKids (L1/L2)        & 181 (85/96)  & $\approx$ 64K (29K/35K) & 109.18(42.27/66.91) \\
% MyST        & --  & $\approx 77k$ & 171.2 \\ 
% \textbf{Total}   & \textbf{--} & \textbf{$\approx$ 141k} & \textbf{280.4} \\
% \midrule
% \multicolumn{4}{l}{\textit{Test data}} \\
% \midrule
% AraKids (L1/L2)       & 66 (26/40)  & 2K (1K/1K)  & 4 (1.57/2.43) \\
% MyST        & --  & 13,180  & 28.32 \\
% \textbf{Total}   & \textbf{--}  & \textbf{15,180}  & \textbf{32.32} \\
% \bottomrule
% \end{tabular}}
% \caption{Child adaptation data overview. The AraKids training set spans ages 7--18, while the test set used in this paper is restricted to ages 7--12 and region-balanced by sampling 250 utterances per region across 8 regions/countries. For MyST, we use the official split.}
% \label{tab:arakids_overview}
% \end{table}

% \subsection{Datasets}
% \label{sec:child_data}

We use AraYoungVoices \cite{chowdhury2026arayoungvoices} and MyST \cite{pradhan-etal-2024} as our primary datasets for adapting, and testing. Hyperparameters are selected on the development sets, and child ASR performance is reported on held-out test sets.

\paragraph{AraYoungVoices}
AraYoungVoices is an Arabic children and adolescent read-speech corpus with 61,574 training utterances from 199 speakers, totaling 102,94 hours. It contains two subsets: \textit{L1}, covering native Arabic speakers from Egyptian, Gulf, Levantine, and North African regions, and \textit{L2}, covering non-native Arabic speakers from Africa, Asia, the Americas, and Europe with at least 13 unique speakers from each reported group. Prompts are educator-designed and fully diacritized. L2 prompts are adapted for different proficiency levels from government-certified books, while controlling lexical and syntactic complexity. 
The AraYoungVoices training set includes speakers aged 7--18 years, while the test set used (AraKids)  includes children aged 7--12 years.
AraKids is released with two reference types: a \emph{prompt
transcription} (\ the written sentence the child was asked to read), and a
\emph{verbatim transcription} produced by human annotators, which records hesitations, repetitions, fillers, and dialectal spellings in place of Modern Standard Arabic forms. We score against the prompt transcription throughout,
so the reported Arabic child WER measures agreement with the prompt transcription rather than with a verbatim transcription.
Table~\ref{tab:arakids_train_regions} reports the number of speakers and utterances in each of the eight L1/L2 groups, for both the training and the test split.
% \paragraph*{AraKids splits}
The AraKids test splits are speaker- and prompt-disjoint (USUP: Unseen Speaker Unseen Prompt), which means no prompt/speaker overlap between train and test. The test set is group-balanced, with approximately 250 utterances sampled from each of the 8 groups: 1,000 utterances for AraKidsL1 and 1,000 for AraKidsL2 yielding a total of 2000 test utterances from 66 speakers of 3.99 hours\footnote{The average utterance duration is 5.64s for AraKidsL1 and 8.74s for AraKidsL2.}. The test set is balanced by utterance count across the eight reported L1/L2 groups; speaker counts are not necessarily equal across groups.
% (seeTable~\ref{tab:arakids_overview}). 

\paragraph{MyST}
We use MyST \cite{pradhan-etal-2024}, an English child-speech corpus of conversational tutoring dialogues, following its official train, development, and test splits. MyST includes students from Grades 3, 4, and 5; the source paper reports grade levels rather than an exact age range.

% \subsection{Adult retention anchors}
% \label{sec:adult_data}
\paragraph{Adult Test sets}
Adult retention is evaluated using Arabic MGB-2 \cite{ali2016mgb} and English LibriSpeech test-clean (Libri-C) \cite{panayotov2015librispeech}. These adult datasets are excluded from child adaptation and are used only to measure whether adapted models preserve performance on adult speech.

\section{Methodology}
\label{sec:method}

We study child ASR adaptation under an adult-retention constraint. Given an adult-pretrained ASR model, the goal is to improve recognition on child speech while preserving the model's original performance on adult speech. We compare three adaptation paradigms: full fine-tuning, parameter-efficient fine-tuning with LoRA, and post-hoc weight-space merging.

Full fine-tuning and LoRA directly adapt the base model using child-speech data. In contrast, weight-space merging is applied after the adaptation where it combines child-adapted checkpoints with the original model to obtain a single model that balances child-speech improvement and adult-speech retention. Since full fine-tuning and LoRA follow standard training procedures, their implementation details are given in Section~\ref{sec:train_decode}. In this section, we focus on the merging methods and the evaluation metrics used to quantify adaptation, retention, and recovery.

\paragraph{\textbf{Post-hoc merging}}
Let $\theta_{\text{base}}$ denote the shared base model and let ${\theta^{1},\ldots,\theta^{N}}$ denote $N$ child-specialized checkpoints that are parameter-compatible with $\theta_{\text{base}}$. Each $\theta^{i}$ is a child-specialized checkpoint produced by adapting the same adult-pretrained base model. Depending on the experiment, the base model is adapted using Arabic child speech, English child speech, or the combined Arabic--English child data through full fine-tuning or LoRA. All merging operations are performed only within a single architecture family (encoder--decoder or encoder--CTC among others) and between checkpoints sharing the same initialization.

\paragraph*{\textbf{Linear interpolation (LERP)}}
We apply linear interpolation in weight space, which takes a weighted average of model parameters. Given the merge set $\{\theta^{1},\ldots,\theta^{N}\}$ (which includes the adult base $\theta_{base}$), the merged model is: $ \theta_m = \sum_{i=1}^{N} \alpha_i\, \theta^{i}$, $\alpha_i \in [0,1]$ $\sum_{i=1}^{N} \alpha_i = 1 .$
% \begin{equation*}
% \footnotesize
%   \theta_m = \sum_{i=1}^{N} \alpha_i\, \theta^{i},
%   \qquad \alpha_i \in [0,1],\ \ \sum_{i=1}^{N} \alpha_i = 1 .
%   \end{equation*}
Note, that unlike task-arithmetic methods, LERP operates directly on the raw weights and does not construct task vectors relative to $\theta_{base}$. 
% We sweep the $\alpha_i$ on development data and select the best-performing configuration.

\paragraph*{\textbf{TIES}}
For each child-adapted checkpoint $\theta^{i}$, we define its \emph{task vector} as $\tau^{i}=\theta^{i}-\theta_{base}$, that is, the parameter update introduced by child adaptation relative to the shared adult base model.
TIES \cite{yadav2023ties} first measures how each child-adapted model differs from the shared adult base model. It removes very small parameter changes, which are assumed to have limited importance. When different adapted models change the same parameter in conflicting directions, TIES selects the direction supported by the majority of models. It then averages only the updates that agree with this direction and adds the resulting update to the base model. In contrast to LERP, which averages raw weights, TIES operates on task vectors and resolves sign conflicts before averaging.
% We tune a small set of TIES hyperparameters on development data and report the selected configuration.
% Sweep results are shown in the plot D in the Figure  ~\ref{fig:merging-sweeps}.
\paragraph{\textbf{Selective merging for encoder--decoder models}}
We also explore selective merging for encoder--decoder ASR models.
The goal is to control which components are merged in order to improve child adaptation and adult retention.
In addition to full-model merging, we evaluate an encoder-only variant.
We merge only the encoder parameters and keep the decoder fixed to the base (adult) checkpoint.
This choice is motivated to analyze the contribution of the encoder and decoder components in model merging. In encoder–decoder ASR architectures, the encoder primarily learns acoustic representations, while the decoder models output token dependencies conditioned on these representations. By keeping the decoder fixed to the adult checkpoint, we isolate the effect of merging updates in the encoder and evaluate whether most of the beneficial adaptation is concentrated in the acoustic representation space. This setup also allows us to assess the impact of restricting merging to a subset of parameters, which may reduce interference between task-specific updates.

\paragraph{\textbf{CTC head anchoring for encoder--CTC models}}
For encoder--CTC models, we merge encoder parameters but keep the adult CTC projection head fixed.
We also tested swapping child-specialized CTC heads trained on  individual AraYoungVoices, MyST, and  joint AraYoungVoices+MyST.

% \begin{figure*}[t]
%   \centering
%   \includegraphics[width=\linewidth]{figures/sweeps_3plus1.png}
%   \caption{Average WER as a function of the interpolation/merging
%   parameter $\alpha$ across decoding setups.\\
%   \textbf{(A)} ED: LERP Encoder-only $\alpha$ sweep on DEV/TEST;\\
%   \textbf{(B)} ED: LERP Encoder Decoder $\alpha$ sweep on DEV/TEST;\\
%   \textbf{(C)} ECTC: CTC head fixed to adult, LERP $\alpha$ sweep on encoder only on DEV/TEST;\\
%   \textbf{(D)} ECTC: TIES parameter sweep (Avg WER) on DEV/TEST. Weights and densities are reported in fixed order [A, B, C], where $A=\theta_{child}^{En}$, $B=\theta_{child}^{Ar}$, and $C=\theta_{child}^{Ar+En}$.\\
%   Blue: DEV; orange: TEST; stars mark the DEV-selected configuration.}
%   \label{fig:merging-sweeps}
% \end{figure*}

% \begin{figure*}[t]
%   \centering
%   \includegraphics[width=\linewidth]{figures/ckpt_sweep_avg_wer.png}
%   \caption{Average WER as a function of the merging
%   parameter $\alpha$ across TIES setups.\\
%   \textbf{ECTC:} TIES parameter sweep (Avg WER) on DEV/TEST. Weights and densities are reported in fixed order [A, B, C], where $A=\theta_{child}^{En}$, $B=\theta_{child}^{Ar}$, and $C=\theta_{child}^{Ar+En}$.\\
%   Blue: DEV; orange: TEST; stars mark the DEV-selected configuration.}
%   \label{fig:merging-sweeps}
% \end{figure*}

\section{Experimental Setup}
\label{sec:exp_setup}

\subsection{ASR Model Architectures}
\label{sec:arch}

We evaluate child ASR adaptation across four ASR families: bilingual encoder--decoder, multilingual Whisper, bilingual encoder--CTC, and AudioLLM-based ASR.

\paragraph*{Bilingual Encoder--Decoder Model (ED-Bi)}
ED-Bi is an adult-pretrained Arabic--English Conformer--Transformer ASR model \cite{chowdhury2021towards}. It uses a self-supervised speech front-end, a 12-block Conformer encoder, and a 6-block Transformer decoder. The model follows a hybrid CTC--attention objective with \(\lambda_{\mathrm{CTC}}=0.3\).

\paragraph*{Whisper Model (ED-Ml)}
ED-Ml is Whisper Large v3, a multilingual encoder--decoder ASR model pretrained on large-scale multilingual and multitask speech data \cite{pmlr-v202-radford23a}. We use it as a high-capacity reference model for adaptation and retention analysis.

\paragraph*{Bilingual Encoder--CTC Model (E-CTC-Bi)}
E-CTC-Bi is an encoder-only Arabic--English ASR model \cite{abbas2026fanar} based on OWSM-CTC \cite{peng2024owsmctcopenencoderonlyspeech}. It uses a frozen Arabic-centric self-supervised front-end \cite{sukhadia2026harness}, Transformer encoder layers, and self-conditioned CTC with intermediate and final-layer CTC losses.

\paragraph*{AudioLLM-based ASR Model (ALLM-ASR-Bi)}
ALLM-ASR-Bi uses \textit{Qwen2.5-Omni-Ar-En-7B} \cite{bhatti2026harmonizing}\footnote{Model obtained upon request from the authors.}, an Arabic--English ASR-adapted variant of Qwen2.5-Omni-7B \cite{qwen2025omni}. It combines audio understanding with instruction-style decoding. All child-adaptation experiments start from this bilingual ASR-oriented checkpoint.
% , adapted using public Arabic and English ASR data, including MGB-3 and QASR \cite{ali2017speechrecognitionchallengewild,mubarak-etal-2021-qasr}.

\subsection{Child Adaptation Configuration}
\label{sec:train_decode}

We compare full fine-tuning, LoRA adaptation, and post-hoc model merging. All checkpoints are selected using development-set WER on AraYoungVoices and MyST.% (see Appendix \ref{} for details).

\paragraph{Full fine-tuning.}
ED-Bi and E-CTC-Bi are adapted for 5 epochs, while ED-Ml is adapted for 1 epoch to reduce overfitting. For E-CTC-Bi, we use AdamW with peak learning rate \(2.5\times10^{-4}\), piecewise linear warm-up, and intermediate CTC losses at layers \(\{6,9,11\}\); the SSL front-end is frozen. For ED-Bi, we use the hybrid CTC--attention objective, Adam with learning rate \(2\times10^{-4}\), and 15k warm-up steps; the XLSR-53 front-end is frozen. For ED-Ml, we use learning rate \(10^{-5}\), and 500 warm-up steps. %, and FP16 mixed precision.

\paragraph{LoRA adaptation.}
LoRA keeps the base model frozen and updates only low-rank adapters. For ED-Bi and E-CTC-Bi, adapters are inserted into attention projection layers with rank \(r=16\), \(\alpha=16\), and dropout 0.05, and optimized for 5 epochs with AdamW at learning rate \(10^{-3}\). For ALLM-ASR-Bi, LoRA starts from the bilingual ASR checkpoint, uses rank \(r=16\), \(\alpha=32\), dropout 0.05, and targets \texttt{all-linear} modules with bf16 training, cosine scheduling, learning rate \(10^{-4}\), and warm-up ratio 0.05.

\paragraph{Model merging.}
Model merging combines the adult base model \(\theta_{base}\) with child-adapted checkpoints obtained through full fine-tuning \(\theta_{child}^{ft}\) or LoRA adaptation \(\theta_{child}^{lora}\). We consider both Arabic-only and Arabic--English child-adapted checkpoints. For ALLM-ASR-Bi, the merge pool additionally includes the original Qwen2.5-Omni-7B checkpoint as a multilingual prior. For E-CTC-Bi, we retain the adult CTC output head after merging, since it gives lower validation WER than the child-adapted CTC heads from either the Arabic-only or Arabic--English checkpoints (Figure ~\ref{fig:ties_ctc_swap}).
For LERP, we sweep \(\alpha \in [0.2,0.8]\) and select \(\alpha_1=0.2, \;\alpha_2 = \alpha_3=0.4\) based on validation-set WER. For TIES, we similarly tune the merge weight \(w\) and density \(d\) on the validation set. The selected merged models are then evaluated on both child and adult benchmarks to quantify the adaptation--retention trade-off. Figures~\ref{fig:ties_wd_sweep} and \ref{fig:ties_ctc_swap} in the Appendix show the merge-parameter sweeps.
%ties_wd_sweep ties_ctc_sweep
% \noindent\textbf{Model merging.}
% Model merging combines the adult base model \(\theta_{base}\) with child-adapted checkpoints from full fine-tuning \(\theta_{child}^{ft}\) or LoRA \(\theta_{child}^{lora}\). We use both Arabic-only and Arabic--English child-adapted checkpoints. For ALLM-ASR-Bi, we additionally include the original Qwen2.5-Omni-7B checkpoint as a multilingual prior. For E-CTC-Bi, we opt for adult CTC-head as it outperfoms the child (Ar or Ar+En) CTC on the validation-set.
% For LERP, we tune \(\alpha \in [0.2,0.8]\) and choose $\alpha=0.2$; for TIES, we tune the merge weight \(w\) and density \(d\). All merge parameters are selected using validation-set WER, and the final models are evaluated on both child and adult benchmarks. Figure~\ref{fig:merging-sweeps}, in Appendix, shows the merge-parameter sweeps.

\subsection{Evaluation Measures}
We use word error rate, \(\mathrm{WER}(\theta,D_*)\), as the primary ASR metric. Before scoring, we apply lightweight normalization to both references and hypotheses: Arabic diacritics are removed, common Arabic letter variants are normalized, including \textit{alef}, \textit{hamza}, \textit{alef maksura} \(\rightarrow\) \textit{ya}, and \textit{teh marbuta}, and Latin text is lowercased. For MyST, we exclude segments with empty references, non-speech-only markers such as \texttt{<silence>} and \texttt{<noise>}, or unintelligible placeholders such as \texttt{(())}. Empty hypotheses are retained when the reference contains speech, since they reflect recognition errors. This reduces scoring artifacts from orthographic variation and non-speech regions.

% We use word error rate, $\mathcal{WER(\theta, D_*)}$, as the primary metric to measure ASR ($\theta$) performances on datasets ($D_*$). To ensure fair scoring under Arabic orthographic variability, we apply a lightweight normalization to both references and hypotheses before computing WER: we remove diacritics, normalize common letter variants (e.g., \textit{alef} forms, \textit{hamza} variants, \textit{alef maksura} $\rightarrow$ \textit{ya}, and \textit{teh marbuta}), and lowercase Latin text.
% For MyST, we additionally exclude from WER computation segments whose reference is empty or contains only non-speech markers (e.g., \texttt{<silence>}, \texttt{<noise>}, \texttt{(silence)}), as well as unintelligible placeholders (e.g., \texttt{(())}); segments with empty hypotheses are also skipped.
% This prevents the metric from penalizing models for orthographic conventions or no-speech regions.

WER remains the primary evaluation metric and is reported for all child- and adult-speech benchmarks. The measures below do not replace WER; they summarize different aspects of the adaptation--retention trade-off and are computed directly from the reported WER values.
Following this, we quantify the trade-offs inherent in child ASR adaptation using three complementary metrics. These metrics separate adult-speech retention, child-speech improvement, and the extent to which merged models recover the gains of child-specific fine-tuning:
\begin{itemize} 
    \item \textit{Retention Index} (\(\mathcal{RI}\)), which measures how well an adapted or merged model preserves adult ASR performance relative to the base model.
    \item \textit{Child Adaptation Gain} (\(\mathcal{AG}\)), which measures how much an adapted model improves over the base model on child speech.
    \item \textit{Adaptation Recovery} (\(\mathcal{AR}\)), which measures how close a merged model is to the child-specific fine-tuned model.
\end{itemize}
For each metric, we also report the \textit{signed WER-point change}, where positive values indicate improvement, while negative values indicate degradation relative to the reference model used by that metric.

\paragraph{Retention Index}
The Retention Index measures how much adult-speech performance is preserved after adaptation:
\begin{equation}
\footnotesize
% \scriptsize
\mathcal{RI}(\theta)=\frac{\mathrm{WER}(\theta_{base}, D_{adult}^{L})}
{\mathrm{WER}(\theta, D_{adult}^{L})}.
\end{equation}
Here, \(L \in \{\mathrm{Ar}, \mathrm{En}\}\), \(\theta_{base}\) is the original adult ASR model, and \(\theta\) is the adapted or merged model.

A value of \(\mathcal{RI}=1.0\) indicates that adult performance is preserved. Values $<$1.0 indicate forgetting, while values  $>$1.0 indicate improved adult performance. To make the magnitude of retention gain or loss explicit, we also report the signed WER-point:
\begin{equation}
\footnotesize
\Delta_{adult}(\theta)=
\mathrm{WER}(\theta_{base}, D_{adult}^{L})-\mathrm{WER}(\theta, D_{adult}^{L}).
\end{equation}
Positive values indicate adult WER improvement, while negative values indicate adult WER degradation. 

\paragraph{Child Adaptation Gain}
Child Adaptation Gain measures the relative WER reduction of an adapted model over the base model on child speech:
\begin{equation}
\footnotesize
\mathcal{AG}(\theta)=
% 100 \times
\frac{\mathrm{WER}(\theta_{base}, D_{child}^{L})-\mathrm{WER}(\theta, D_{child}^{L})}{\mathrm{WER}(\theta_{base}, D_{child}^{L})}.
\end{equation}
A value of \(\mathcal{AG}=0\) means that the adapted model performs the same as the base model on child speech. Positive values indicate improvement over the base model, while negative values indicate degradation.

The absolute magnitude of the gain or loss explicit are reported using:
\begin{equation}
\footnotesize
\Delta_{child}(\theta)=
\mathrm{WER}(\theta_{base}, D_{child}^{L})
-
\mathrm{WER}(\theta, D_{child}^{L}).
\end{equation}
Positive values indicate improvement, while negative values indicate that adaptation hurts child ASR.

\paragraph{Adaptation Recovery}
For merged models, Adaptation Recovery measures how much of the child-specific adaptation gain is recovered by the merged model:
\begin{equation}
\footnotesize
\mathcal{AR}(\theta_m)=
% 100 \times
\frac{\mathrm{WER}(\theta_{base}, D_{child}^{L})-\mathrm{WER}(\theta_m, D_{child}^{L})}
{\mathrm{WER}(\theta_{base}, D_{child}^{L})-\mathrm{WER}(\theta_{ft}, D_{child}^{L})}.
\end{equation}

Here, \(\theta_{ft}\) is the best child-specific adapted model, obtained either through full fine-tuning or LoRA adaptation, and \(\theta_m\) is the merged model. This metric is computed when \(\theta_{ft}\) improves over \(\theta_{base}\) on child speech.

A value of \(\mathcal{AR}=1\) means that the merged model matches the child-specific adapted model. Values with in $[0, \; 1)$ indicate partial recovery of the child-specific adaptation gain, and above 1 indicate that the merged model surpasses the child-specific adapted model, and values $<0$ indicate that the merged model performs worse than the base model on child speech.
We also report the remaining WER gap to the child-specific adapted model:
\begin{equation}
\footnotesize
\Delta_{rec}(\theta_m)=
\mathrm{WER}(\theta_{ft}, D_{child}^{L})-\mathrm{WER}(\theta_m, D_{child}^{L}).
\end{equation}
Positive values indicate that the merged model outperforms the child-specific adapted model, while negative values indicate that it remains worse. 

\begin{table*}[!ht]
\centering
\scriptsize
\setlength{\tabcolsep}{2.5pt}
\renewcommand{\arraystretch}{1.02}
\resizebox{\textwidth}{!}{%
\begin{tabular}{@{}lrrrrrr@{\hspace{2.5em}}lrrrrrr@{}}
\toprule
\multicolumn{7}{c}{\textit{Encoder--Decoder (ED-Bi)}} &
\multicolumn{7}{c}{\textit{Encoder--CTC (E-CTC-Bi)}} \\
\cmidrule(r{2.5em}){1-7}
\cmidrule(l{2.5em}){8-14}
\textbf{Model} & \textbf{ArL1} & \textbf{ArL2} & \textbf{MyST} & \textbf{Libri-C} & \textbf{MGB2} & \textbf{Avg.} &
\textbf{Model} & \textbf{ArL1} & \textbf{ArL2} & \textbf{MyST} & \textbf{Libri-C} & \textbf{MGB2} & \textbf{Avg.} \\
\midrule
$\theta_{base}$              & 9.14  & 44.11  & 36.68 & 9.86 & 10.92 & 22.142 &
$\theta_{base}$              & 14.72 & 63.37  & 26.63 & \textcolor{blue}{3.83} & \textcolor{blue}{13.18} & 24.346 \\
\addlinespace[1pt]
$\theta_{child}^{Ar}(ft)$        & 6.41  & 26.76  & 44.23 & 12.21 & \textcolor{blue}{10.70} & 20.062 &
$\theta_{child}^{Ar}(ft)$        & 13.11 & \textcolor{blue}{16.32} & 77.32 & 18.64 & 23.63 & 29.804 \\
\addlinespace[1pt]
$\theta_{child}^{En}(ft)$        & 25.75 & 106.98 & 22.68 & 6.66 & 14.16 & 35.246 &
$\theta_{child}^{En}(ft)$        & 51.44 & 92.73  & \textcolor{blue}{15.11} & 6.80 & 26.26 & 38.468 \\
\addlinespace[1pt]
$\theta_{child}^{Ar+En}(ft)$     & \textcolor{blue}{6.37} & \textcolor{blue}{22.71} & \textcolor{blue}{22.52} & \textcolor{blue}{6.66} & 12.97 & \textcolor{blue}{14.246} &
$\theta_{child}^{Ar+En}(ft)$     & \textcolor{blue}{12.45} & 24.24 & 15.90 & 6.66 & 21.20 & \textcolor{blue}{16.090} \\
\addlinespace[1pt]
$\theta_{child}^{Ar+En}(l)$ & 7.16 & 26.49 & 23.49 & 7.12 & 11.99 & 15.250 &
$\theta_{child}^{Ar+En}(l)$ & 15.16 & 37.65 & 18.57 & 5.91 & 16.77 & 18.810 \\
\midrule
\multicolumn{7}{c}{\textit{Whisper Large v3 (ED-Ml)}} &
\multicolumn{7}{c}{\textit{Qwen2.5--Omni--7B--Ar-En (ALLM-ASR-Bi)}} \\
\cmidrule(r{2.5em}){1-7}
\cmidrule(l{2.5em}){8-14}
\textbf{Model} & \textbf{ArL1} & \textbf{ArL2} & \textbf{MyST} & \textbf{Libri-C} & \textbf{MGB2} & \textbf{Avg.} &
\textbf{Model} & \textbf{ArL1} & \textbf{ArL2} & \textbf{MyST} & \textbf{Libri-C} & \textbf{MGB2} & \textbf{Avg.} \\
\midrule
 & & & & & & &
Qwen2.5O & 24.24 & 73.53 & 15.64 & 4.07 & 34.24 & 30.344 \\
\addlinespace[1pt]
$\theta_{base}$ & 9.68 & 40.68 & 21.96 & \textcolor{blue}{2.47} & \textcolor{blue}{15.15} & 17.988 &
$\theta_{base}$ & 6.96 & 34.58 & 35.87 & \textcolor{blue}{1.48} & \textcolor{blue}{13.27} & 18.432 \\
\addlinespace[1pt]
$\theta_{child}^{Ar+En}(ft)$ & \textcolor{blue}{8.70} & \textcolor{blue}{11.84} & \textcolor{blue}{14.47} & 3.11 & 23.52 & \textcolor{blue}{12.328} &
$\theta_{child}^{Ar+En}(l)$ & \textcolor{blue}{6.50} & \textcolor{blue}{26.72} & \textcolor{blue}{15.55} & 1.56 & 16.83 & \textcolor{blue}{13.432} \\
\bottomrule
\end{tabular}%
}
% \vspace{-0.1cm}
\caption{Reported WER (\%) for the adult $\theta_{base}$ model  and adapted ASR models across four architectures. ArL1: AraKidsL1, ArL2: AraKidsL2, ft: full fine-tuning, l: LoRA and Qwen2.5: Base weights for the Qwen Omni model \cite{qwen2025omni}. Blue marks the best score within each architecture block. Lower WER is better. }
% \vspace{-0.2cm}
\label{tab:child_asr_wer}
\end{table*}

% \begin{table}[t]
% \centering
% \scalebox{0.55}{
% \setlength{\tabcolsep}{4pt}
% \renewcommand{\arraystretch}{0.85}
% \begin{tabular}{lcccccc}
\begin{table}[t]
\centering
\scriptsize
\setlength{\tabcolsep}{2pt}
\renewcommand{\arraystretch}{0.85}
\resizebox{\columnwidth}{!}{%
\begin{tabular}{lcccccc}
\toprule
\textbf{MOp} & \textbf{ArL1} & \textbf{ArL2} & \textbf{MyST} & \textbf{Libri-C} & \textbf{MGB2} & \textbf{Avg. WER} \\
\midrule
\multicolumn{7}{c}{\textit{Encoder--Decoder (ED-Bi)}} \\ \midrule
LERP          & \textcolor{blue}{7.12} & 27.01 & 30.43 & 09.06 & 11.00 & 16.924 \\
LERP (Enc)  & 7.22 & 27.38 & 31.26 & 8.65 & 11.00 & 17.102 \\
TIES          & 7.14 & \textcolor{blue}{26.76} & 28.67 & 9.23 & 11.26 & 16.612 \\
TIES (Enc)  & 7.35 & 27.01 & \textcolor{blue}{28.25} & \textcolor{blue}{8.16} & \textcolor{blue}{10.86} & \textcolor{blue}{16.326} \\
\midrule
\multicolumn{7}{c}{\textit{Encoder--CTC (E-CTC-Bi)}} \\ \midrule
LERP         & 11.20 & 23.08 & 20.25 & \textcolor{blue}{4.37} & \textcolor{blue}{14.63} & 14.706 \\
TIES         & \textcolor{blue}{11.07} & \textcolor{blue}{18.94} & \textcolor{blue}{18.68} & 5.66 & 18.71 & \textcolor{blue}{14.612} \\
\midrule
\multicolumn{7}{c}{\textit{Whisper-Large v3 (ED-Ml)}} \\ \midrule
LERP          & \textcolor{blue}{6.55} & 20.01 & 17.78 & 2.43 & \textcolor{blue}{14.49} & 12.252 \\
LERP (Enc)  & 7.64 & 26.38 & 20.45 & \textcolor{blue}{2.41} & 15.00 & 14.376 \\
TIES          & 8.48 & \textcolor{blue}{11.37} & \textcolor{blue}{14.36} & 2.91 & 22.68 & \textcolor{blue}{11.96} \\
TIES (Enc)  & 7.16 & 19.83 & 17.55 & 2.61 & 17.01 & 12.832 \\
\midrule
\multicolumn{7}{c}{\textit{Qwen2.5-Omni-7B (ALLM-ASR-Bi)}} \\ \midrule
LERP & 6.47 & \textcolor{blue}{27.01} & 12.79 & \textcolor{blue}{1.59} & 14.33 & 12.438 \\
TIES & \textcolor{blue}{6.44} & 27.15 & \textcolor{blue}{12.78} & 1.60 & \textcolor{blue}{14.19} & \textcolor{blue}{12.432} \\
\bottomrule
\end{tabular}}
\caption{ Reported WER (\%) for different merge operations. Best scores within each architecture are highlighted in blue. MOp: merge operation.}
\vspace{-0.2cm}
\label{tab:mergeops_wer}
\end{table}

\section{Results}
\label{sec:results}

\paragraph{Child Adaptation and Adult Retention with Fine-tuning}
% \label{sec:results_adaptation}

Table~\ref{tab:child_asr_wer} reports WER for the adult base models and child-adapted checkpoints. Across architectures, child adaptation improves recognition on child speech, with the largest gains seen on ArL2 and MyST. For ED-Bi, bilingual child adaptation reduces WER by 49\% and 38.6\% on ArL2 and MyST respectively, giving the best average WER within the ED block. For ED-Ml, bilingual adaptation gives even larger gains on ArL2. For ALLM-ASR-Bi, child LoRA improves over all child benchmarks relative to the bilingual adult base, most notably the LoRA is able to recover MyST, where WER drops from 35.87 to 15.55.

The results also show that language-specific adaptation can be unstable. In E-CTC-Bi, Arabic-only adaptation gives the best ArL2 WER, but severely degrades MyST. English-only adaptation gives the best MyST WER, but substantially hurts Arabic child performance. Bilingual adaptation is therefore more balanced, giving the best average WER for both ED-Bi and E-CTC-Bi.

However, child adaptation is not retention-neutral. Adult performance often degrades after child adaptation, especially on MGB-2. For example, MGB-2 WER increases for all architectures under bilingual child adaptation. This confirms the need to evaluate child ASR adaptation together with adult-speech retention.

\paragraph{Merging and Adaptation--Retention Trade-offs}
% \label{sec:results_tradeoffs}

Tables~\ref{tab:mergeops_wer} and \ref{tab:ri_ar_tradeoffs} (along with Table \ref{tab:ag_tradeoffs} in Appendix) show that weight-space merging often improves the adaptation--retention balance, but its effect depends on the architecture and merge operation. Direct child adaptation usually gives strong child-speech gains, but it can also reduce adult ASR performance, especially on Arabic. Merging helps to recover child gain from fine-tuning while reducing the adult-retention cost.

For \textit{ED-Bi}, merging gives modest benefits. The best merged variant is encoder-level TIES, with the lowest average WER for the ED-Bi. However, merged models do not fully match the child WER of bilingual fine-tuning. Their main advantage is retention. LERP keeps MGB-2 close to the base model with \(\mathcal{RI}=0.99\) and only \(-0.08\) WER loss, while TIES gives \(\mathcal{RI}=0.97\). Thus, for ED-Bi, merging is useful for retention, but direct bilingual fine-tuning remains stronger for child WER.

For \textit{E-CTC-Bi}, merging provides a clearer trade-off improvement. Full fine-tuning gives strong child gains but causes substantial adult forgetting, with \(\mathcal{RI}=0.58\) on Libri-C and \(\mathcal{RI}=0.62\) on MGB-2. LERP improves retention to \(\mathcal{RI}=0.88\) and \(0.90\), respectively, while still improving child WER. TIES gives the strongest child recovery, exceeding the child-specific reference on ArL1 and ArL2, but with weaker adult retention than LERP. This shows that LERP is more retention-preserving, whereas TIES is more child-adaptation-oriented.

For \textit{ED-Ml}, the same pattern holds. LERP is the safest merge operation for adult retention, reaching \(\mathcal{RI}>1.0\) on both Libri-C and MGB-2, while still improving child WER over the base model. TIES gives stronger child recovery, matching or slightly surpassing the child-specific reference on ArL2 and MyST, but it incurs a larger MGB-2 retention cost.

For \textit{ALLM-ASR-Bi}, merging gives the strongest overall benefit. Both LERP and TIES improve over the child LoRA checkpoint, and TIES gives the best average WER. It improves MyST WER from 15.55 to 12.78 and reduces MGB-2 WER from 16.83 to 14.19. In the trade-off metrics, TIES also gives the best MyST adaptation gain, \(\mathcal{AG}=\)0.644(a 64.4\% relative WER reduction), and improves MGB-2 retention to \(\mathcal{RI}=0.94\).

\paragraph{Trends} Overall, the results show three trends. First, child ASR adaptation is necessary, especially for non-native Arabic children (ArL2) and English (MyST) data in bilingual modeling setup. Second, direct child adaptation can substantially reduce adult ASR performance, making retention-aware evaluation essential. Third, weight-space merging often provides a better compromise, especially for E-CTC-Bi, ED-Ml, and ALLM-ASR-Bi. LERP generally favors adult retention, while TIES more aggressively recovers child-speech gains and can, in some settings, improve both child and adult ASR, as reported above for ALLM-ASR-Bi. This recurring pattern across multiple ASR architectures is one of the central findings of this study.

\begin{table}[!t]
\centering
\scriptsize
\setlength{\tabcolsep}{3pt}
\renewcommand{\arraystretch}{1.15}
\resizebox{\columnwidth}{!}{%
\begin{tabular}{@{}l cc ccc@{}}
\toprule
& \multicolumn{2}{c}{\textbf{$\mathcal{RI}$} \scriptsize($\Delta_{adult}$)} 
& \multicolumn{3}{c}{\textbf{$\mathcal{AR}$} \scriptsize($\Delta_{rec}$)} \\
\cmidrule(lr){2-3}\cmidrule(lr){4-6}
\textbf{Model} & \textbf{Libri-C} & \textbf{MGB2} 
& \textbf{ArL1} & \textbf{ArL2} & \textbf{MyST} \\
\midrule
\multicolumn{6}{c}{\textit{Encoder--Decoder (ED-Bi)}} \\
\cmidrule(lr){1-6}
\rowcolor{refgray} \textit{FT} 
& \ddB{1.48}{+3.20} & \dd{0.84}{-2.05} 
& \textit{--} & \textit{--} & \textit{--} \\
LoRA 
& \dd{1.38}{+2.74} & \dd{0.91}{-1.07} 
& \dd{0.715}{-0.79} & \ddB{0.823}{-3.78} & \ddB{0.931}{-0.97} \\
LERP 
& \dd{1.09}{+0.80} & \ddB{0.99}{-0.08} 
& \ddB{0.729}{-0.75} & \dd{0.799}{-4.30} & \dd{0.441}{-7.91} \\
TIES 
& \dd{1.07}{+0.63} & \dd{0.97}{-0.34} 
& \dd{0.722}{-0.77} & \dd{0.811}{-4.05} & \dd{0.566}{-6.15} \\
\midrule
\multicolumn{6}{c}{\textit{Encoder--CTC (E-CTC-Bi)}} \\
\cmidrule(lr){1-6}
\rowcolor{refgray} \textit{FT} 
& \dd{0.58}{-2.83} & \dd{0.62}{-8.02} 
& \textit{--} & \textit{--} & \textit{--} \\
LoRA 
& \dd{0.65}{-2.08} & \dd{0.79}{-3.59} 
& \dd{-0.194}{-2.71} & \dd{0.657}{-13.41} & \ddB{0.751}{-2.67} \\
LERP 
& \ddB{0.88}{-0.54} & \ddB{0.90}{-1.45} 
& \dd{1.551}{+1.25} & \dd{1.030}{+1.16} & \dd{0.595}{-4.35} \\
TIES 
& \dd{0.68}{-1.83} & \dd{0.70}{-5.53} 
& \ddB{1.608}{+1.38} & \ddB{1.135}{+5.30} & \dd{0.741}{-2.78} \\
\midrule
\multicolumn{6}{c}{\textit{Whisper Large v3 (ED-Ml)}} \\
\cmidrule(lr){1-6}
\rowcolor{refgray} \textit{FT} 
& \dd{0.79}{-0.64} & \dd{0.64}{-8.37} 
& \textit{--} & \textit{--} & \textit{--} \\
LERP 
& \ddB{1.02}{+0.04} & \ddB{1.05}{+0.66} 
& \ddB{3.194}{+2.15} & \dd{0.717}{-8.17} & \dd{0.558}{-3.31} \\
TIES 
& \dd{0.85}{-0.44} & \dd{0.67}{-7.53} 
& \dd{1.224}{+0.22} & \ddB{1.016}{+0.47} & \ddB{1.015}{+0.11} \\
\midrule
\multicolumn{6}{c}{\textit{Qwen2.5--Omni--7B (ALLM-ASR-Bi)}} \\
\cmidrule(lr){1-6}
\rowcolor{refgray} \textit{LoRA} 
& \ddB{0.95}{-0.08} & \dd{0.79}{-3.56} 
& \textit{--} & \textit{--} & \textit{--} \\
LERP 
& \dd{0.93}{-0.11} & \dd{0.93}{-1.06} 
& \dd{1.065}{+0.03} & \ddB{0.963}{-0.29} & \dd{1.136}{+2.76} \\
TIES 
& \dd{0.92}{-0.12} & \ddB{0.94}{-0.92} 
& \ddB{1.130}{+0.06} & \dd{0.945}{-0.43} & \ddB{1.136}{+2.77} \\
\bottomrule
\end{tabular}}
\caption{Retention and adaptation recovery trade-offs per benchmark.
$\mathcal{RI}$ (adult sets) and $\mathcal{AR}$ (child sets) are reported with their signed WER point change ($\Delta_{adult}$, $\Delta_{rec}$); higher values are better. Reference models are on a gray background, and the best $\Delta$ is in \textcolor{bestblue}{blue}.}

\label{tab:ri_ar_tradeoffs}
\vspace{-0.3cm}
\end{table}

\section{Conclusion}
We presented a retention-aware study of adult-to-child ASR adaptation across encoder--decoder, encoder--CTC, Whisper, and AudioLLM-based ASR systems. Using Arabic native and non-native child speech, English MyST, and adult benchmarks from MGB-2 and LibriSpeech test-clean, we evaluated full fine-tuning, LoRA, and post-hoc weight-space merging with WER, Retention Index, and Adaptation Recovery. The results show that child adaptation is necessary, but direct adaptation can reduce adult ASR performance. Bilingual adaptation is generally more stable than language-specific adaptation. Weight-space merging often improves the adaptation--retention balance, particularly for encoder--CTC, Whisper, and AudioLLM-based ASR models, while direct bilingual fine-tuning remains strongest for the Ar-En encoder--decoder model in raw WER. Overall, our findings highlight model merging as a simple and effective strategy for building ASR systems that better serve both child and adult speakers.

\section{Limitations}
This study focuses on a retention-aware comparison of widely used adaptation strategies rather than proposing a new ASR architecture or merging algorithm. This choice allows us to isolate how full fine-tuning, LoRA, and post-hoc merging behave across different ASR families, but it also leaves room for more specialized adaptation methods, such as continual-learning objectives, rehearsal-based training, or retention-aware regularization.
Our evaluation covers Arabic and English child speech, with Arabic further divided into native and non-native speakers. While this provides a useful bilingual and cross-nativity setting, the results may not fully generalize to other languages, spontaneous child speech, or highly noisy real-world environments. Future work can extend the analysis to broader ranges, more dialects, additional speaking styles, and deployment-oriented conditions such as classrooms, homes, and mobile recordings.
Broad regional categories may also mask country-specific and dialect-specific variation, so these results should not be interpreted as country-level performance estimates.
Speaker-age classification and model routing are valid alternatives, but they address a different system design. Routing requires an additional age classifier and multiple models, which can increase latency, deployment cost, and error propagation, especially when adults and children speak in the same session. A direct comparison between routing and unified adaptation is left for future work.
% We conducted a retention-aware empirical study of adult$\rightarrow$child ASR adaptation across four model families (ED, E-CTC, Whisper, and an AudioLLM-based ASR) and three paradigms (full fine-tuning, LoRA, and post-hoc LERP/TIES merging). We evaluate on AraKidsL1/L2 and MyST, with adult retention anchored on LibriSpeech test-clean and MGB-2, reporting WER plus Retention Index and Adaptation Recovery. Across architectures, we see clear adaptation--retention trade-offs: naive fine-tuning can deliver strong child gains but may trigger cross-language interference (especially E-CTC), while weight-space merging often improves the frontier, and encoder-only merging provides an additional retention knob for encoder--decoder systems.

\section{Acknowledgment}
The work is supported by HBKU flagship research grant
(HBKU-INT-VPR-FRG-03-09). The findings achieved herein
are solely the responsibility of the authors.

\bibliography{custom}

\clearpage
\appendix
\section{Appendix A.}
\label{sec:appendix}
\subsection{Parameter sweep ablation}
\label{sec:ablation}

We ablate the TIES merging configuration along two axes:
the per-expert weights/densities (Fig.~\ref{fig:ties_wd_sweep}) and the
choice of CTC head (Fig.~\ref{fig:ties_ctc_swap}).

% Appendix-style figure numbering (Figure A.1, A.2, ...)
\renewcommand{\thefigure}{A.\arabic{figure}}
\setcounter{figure}{0}

\begin{figure}[!ht]
  \centering
  \includegraphics[width=\columnwidth]{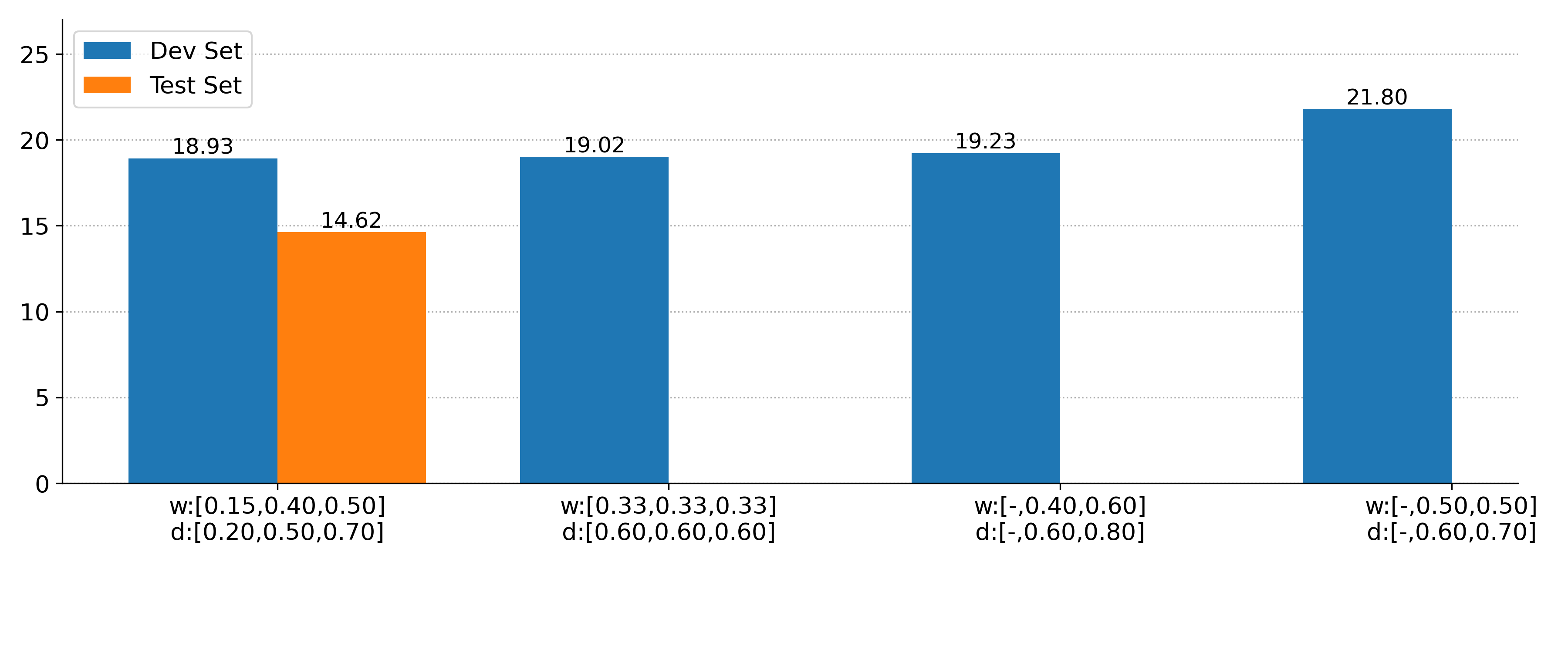}
  \caption{TIES merging weight/density sweep (average WER, \%). Each bar group
  merges the base model $M_0$ with expert checkpoints $A$, $B$, $C$ using
  per-expert weights $w$ and densities $d$, reported in fixed order $[A,B,C]$,
  where $A=\theta_{child}^{En}$, $B=\theta_{child}^{Ar}$, and
  $C=\theta_{child}^{Ar+En}$; all variants retain the adult CTC head. Dev (blue)
  and Test (orange) WER are shown. The selected configuration
  $w{=}[0.15,0.40,0.50]$, $d{=}[0.20,0.50,0.70]$ attains the lowest WER.}
  \label{fig:ties_wd_sweep}
\end{figure}

\begin{figure}[!ht]
  \centering
  \includegraphics[width=\columnwidth]{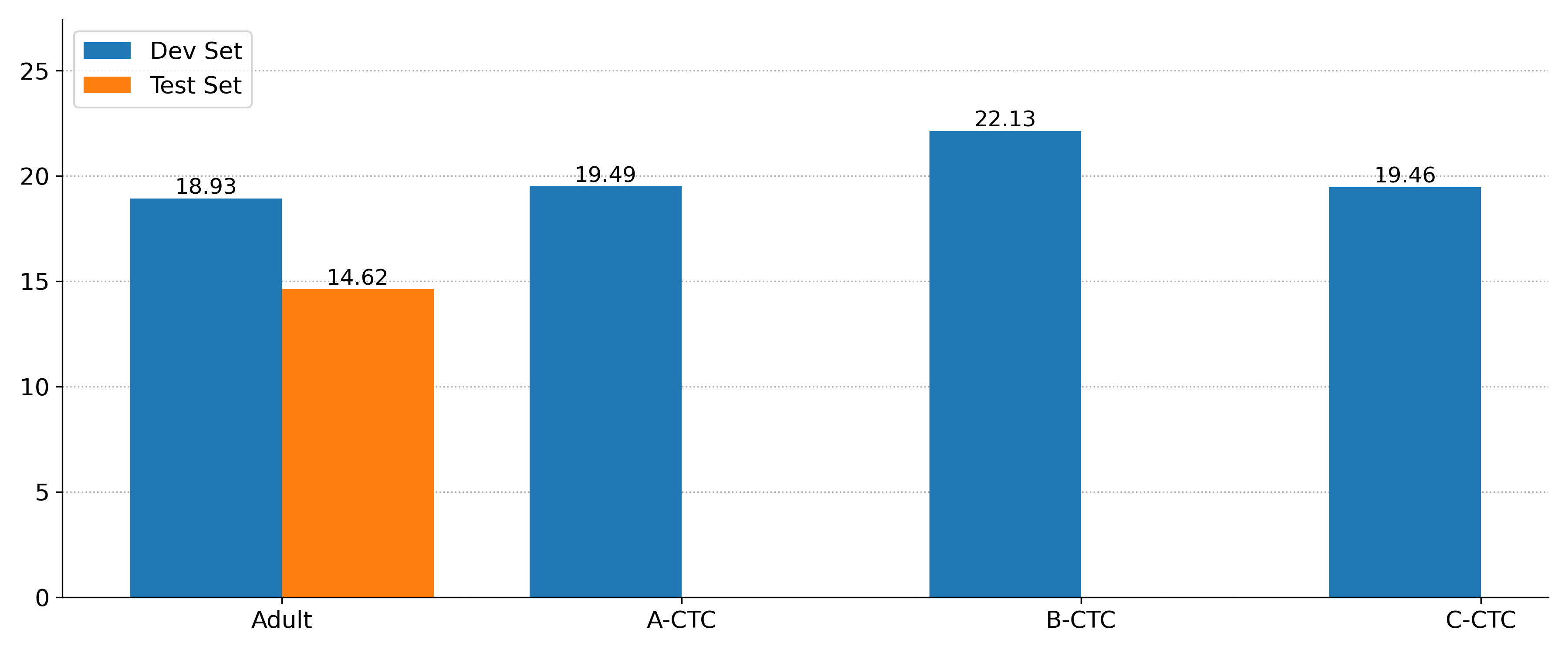}
  \caption{Effect of CTC head swapping on average WER (\%), with the TIES merge
  fixed to the selected configuration ($w{=}[0.15,0.40,0.50]$,
  $d{=}[0.20,0.50,0.70]$). \emph{Adult} uses the base adult CTC head;
  \emph{A-CTC}/\emph{B-CTC}/\emph{C-CTC} substitute the CTC head from expert
  $A$/$B$/$C$ while leaving the merged encoder unchanged. The adult head
  remains best on Dev.}
  \label{fig:ties_ctc_swap}
\end{figure}

\subsection{Adaptation Gain interpretation}
\label{sec:AG}
Table~\ref{tab:ag_tradeoffs} reports Child Adaptation Gain (\(\mathcal{AG}\)) per benchmark, complementing the retention and recovery view in Table~\ref{tab:ri_ar_tradeoffs}. The relative ordering of methods is architecture-dependent. 
For ED, full fine-tuning yields the largest child gains on all three child sets (\(\mathcal{AG}=0.303,\,0.485,\,0.386\) on AraKidsL1, AraKidsL2, and MyST); LoRA and both merge operations give smaller gains, with the gap widest on MyST (\(\mathcal{AG}\) dropping to \(0.170\)--\(0.218\) for merging).

For E-CTC, TIES is strongest, giving the highest \(\mathcal{AG}\) on AraKidsL1
(\(0.248\)) and AraKidsL2 (\(0.701\)) and even exceeding full fine-tuning, while
LoRA is the weakest and slightly degrades AraKidsL1 (\(\mathcal{AG}=-0.030\)).

For Whisper, LERP gives the best AraKidsL1 gain (\(0.323\)), whereas TIES is
strongest on AraKidsL2 (\(0.721\)) and MyST (\(0.346\)). 

For Qwen2.5--Omni--7B,
child gains on the Arabic sets are small across methods, but both LERP and TIES
clearly improve MyST over the LoRA reference (\(\mathcal{AG}\) rising from
\(0.566\) to \(\approx 0.64\)). Overall, \(\mathcal{AG}\) confirms that the
strongest child-speech improvement comes from full fine-tuning for ED, but from
TIES merging for E-CTC and Whisper.
% =========================
% AG TABLE
% =========================
\begin{table}[t]
\centering
\scriptsize
\setlength{\tabcolsep}{3.5pt}
\renewcommand{\arraystretch}{1.15}
\begin{tabular}{@{}lccc@{}}
\toprule
\textbf{Model} &
\multicolumn{3}{c}{\textbf{$\mathcal{AG}$} \scriptsize($\Delta_{child}$)} \\
\cmidrule(lr){2-4}
& \textbf{ArL1} & \textbf{ArL2} & \textbf{MyST} \\
\midrule
\multicolumn{4}{c}{\textit{Encoder--Decoder (ED-Bi)}} \\
\cmidrule(lr){1-4}
\rowcolor{refgray} \textit{FT (ref)} & \ddB{0.303}{+2.77} & \ddB{0.485}{+21.40} & \ddB{0.386}{+14.16} \\
LoRA & \dd{0.217}{+1.98} & \dd{0.399}{+17.62} & \dd{0.360}{+13.19} \\
LERP & \dd{0.221}{+2.02} & \dd{0.388}{+17.10} & \dd{0.170}{+6.25} \\
TIES & \dd{0.219}{+2.00} & \dd{0.393}{+17.35} & \dd{0.218}{+8.01} \\
\midrule
\multicolumn{4}{c}{\textit{Encoder--CTC (E-CTC-Bi)}} \\
\cmidrule(lr){1-4}
\rowcolor{refgray} \textit{FT (ref)} & \dd{0.154}{+2.27} & \dd{0.617}{+39.13} & \ddB{0.403}{+10.73} \\
LoRA & \dd{-0.030}{-0.44} & \dd{0.406}{+25.72} & \dd{0.303}{+8.06} \\
LERP & \dd{0.239}{+3.52} & \dd{0.636}{+40.29} & \dd{0.240}{+6.38} \\
TIES & \ddB{0.248}{+3.65} & \ddB{0.701}{+44.43} & \dd{0.299}{+7.95} \\
\midrule
\multicolumn{4}{c}{\textit{Whisper Large v3 (ED-Ml)}} \\
\cmidrule(lr){1-4}
\rowcolor{refgray} \textit{FT (ref)} & \dd{0.101}{+0.98} & \dd{0.709}{+28.84} & \dd{0.341}{+7.49} \\
LERP & \ddB{0.323}{+3.13} & \dd{0.508}{+20.67} & \dd{0.190}{+4.18} \\
TIES & \dd{0.124}{+1.20} & \ddB{0.721}{+29.31} & \ddB{0.346}{+7.60} \\
\midrule
\multicolumn{4}{c}{\textit{Qwen2.5--Omni--7B (ALLM-ASR-Bi)}} \\
\cmidrule(lr){1-4}
\rowcolor{refgray} \textit{LoRA (ref)} & \dd{0.066}{+0.46} & \ddB{0.227}{+7.86} & \dd{0.566}{+20.32} \\
LERP & \dd{0.070}{+0.49} & \dd{0.219}{+7.57} & \dd{0.643}{+23.08} \\
TIES & \ddB{0.075}{+0.52} & \dd{0.215}{+7.43} & \ddB{0.644}{+23.09} \\
\bottomrule
\end{tabular}
\caption{Child adaptation gain per benchmark. $\mathcal{AG}$ = Child Adaptation Gain on child sets, reported as relative WER reduction with $\Delta_{child}$ in WER points vs.\ $\theta_{base}$. Higher $\mathcal{AG}$ and positive $\Delta$ are better. Within each architecture block, the best $\Delta$ per column is in blue.}
\label{tab:ag_tradeoffs}
\end{table}

\end{document}